\documentclass[10pt,twocolumn,letterpaper]{article}

\usepackage{iccv}
\usepackage{times}
\usepackage{epsfig}
\usepackage{graphicx}
\usepackage{amsmath}
\DeclareMathOperator*{\argmax}{argmax}
\usepackage{amssymb}

\usepackage{algorithm}
\usepackage{algpseudocode}
\usepackage{indentfirst}
\usepackage{authblk}
\usepackage{hyperref}

\iccvfinalcopy

\def\iccvPaperID{****}

\ificcvfinal\fi

\begin{document}

\title{1st Place in ICCV 2023 Workshop Challenge Track 1 on Resource Efficient Deep Learning for Computer Vision: Budgeted Model Training Challenge \vspace{-3.5ex}}

\author[12]{\emph{Youngjun Kwak}}
\author[1]{\emph{Seonghun Jeong}}
\author[1]{\emph{Yunseung Lee}}
\author[2]{\emph{Changick Kim\thanks{Corresponding author}}}

\affil[1]{KakaoBank Corp., South Korea}
\affil[2]{Department of Electrical Engineering, KAIST, South Korea}
\affil[ ]{\small{\texttt{\{vivaan.yjkwak,bentley.j,yun.lee\}@lab.kakaobank.com},
\texttt{\{yjk.kwak,changick\}@kaist.ac.kr}}\vspace{-2.5ex}}

\maketitle

\ificcvfinal\thispagestyle{empty}\fi

\begin{abstract}
The budgeted model training challenge aims to train an efficient classification model under resource limitations. To tackle this task in ImageNet-100, we describe a simple yet effective resource-aware backbone search framework composed of profile and instantiation phases. In addition, we employ multi-resolution ensembles to boost inference accuracy on limited resources. The profile phase obeys time and memory constraints to determine the models' optimal batch-size, max epochs, and automatic mixed precision (AMP). And the instantiation phase trains models with the determined parameters from the profile phase. For improving intra-domain generalizations, the multi-resolution ensembles are formed by two-resolution images with randomly applied flips. We present a comprehensive analysis with expensive experiments. Based on our approach, we win first place in International Conference on Computer Vision (ICCV) 2023 Workshop Challenge Track 1 on \textbf{R}esource Efficient Deep Learning for \textbf{C}omputer \textbf{V}ision (RCV).
\end{abstract}

\vspace{-2.0ex}
\section{Our Approach}
We provide a detailed account of our two victorious approaches: 1) the resource-aware backbone search comprises \textit{profile} and \textit{instantiation} phases with the aim of identifying the optimal models that utilize either automatic mixed precision (AMP) or single precision floating point format (FP32), and 2) the proposed ensemble consists of multi-inferences with randomly flipped multi-resolution images to improve accuracy on the time and memory constraints.

\subsection{Problem Description}
ImageNet-100 is a subset of ImageNet-1K by choosing 100 classes and splitting them into training, validation, and test sets~\cite{rcvt1}. With the benchmark, our model is trained to maximize accuracy on the test set with a constraint of GPU memory (6 GB) and training time (9 GPU hours).

\begin{algorithm}[tb]
\caption{\small Procedure of our resource-aware backbone search}\label{alg:cap1}
\begin{algorithmic}
\scriptsize
\State \emph{\textbf{Profile phase:} Profile candidate models.}

\State Let denote available models $F=\{f_{\theta, q \in \{\text{AMP}, \text{FP32}\}}^{i}\}_{i=0}^{N}$ of PyTorch's timm. Tune parameters such as batch-size $b\in \mathbb{Z}^{+}$, input-size $s^{HW} \in\{160, 224\}$, and max-epochs $m$ by pre-estimated training time $t\in \mathbb{R}^{+}$.
\State Candidates $C=list()$
\For {$f_{\theta, q}^{i} \gets F$}
    \For{$s^{hw} \gets$ $s^{HW}$}
        \State $b^{i}, m^{i} \gets$ PROFILE\_MODEL($f_{\theta, q}^{i}, s^{hw}$)
        \State $C$.push\_back($(f_{\theta, q}^{i}, s^{hw}, b^{i}, m^{i})$)
    \EndFor
\EndFor

\Function{PROFILE\_MODEL}{$f_{\theta, q}^{i}, s^{hw}$}
    \State Calculate the number of learnable-parameters $|f_{\theta, q}^{i}|$ with given $i^{th}$ model.
    \State Find maximum batch-size $b$ with given $|f_{\theta, q}^{i}|$, $s^{hw}$, and 6GB memory by $\phi$:
    \State $\argmax\limits_{b \in \mathbb{Z}^{+}} \phi(b):=\{b\in \mathbb{Z}^{+}:\phi(\hat{b}) < \phi(b) , \forall \text{ } \hat{b} \in \mathbb{Z}^{+}\}$.
    \State $t \gets$ ESTIMATE\_ONE\_EPOCH\_TIME($f_{\theta, q}^{i}, b, s^{hw}$)
    \State Calculate max epochs $m$ with estimated $t$ and given 9 GPU hours.
    \State \Return $b, m$
\EndFunction
\Function{ESTIMATE\_ONE\_EPOCH\_TIME}{$f_{\theta, q}^{i}, b, s^{hw}$}
    \State Define dummy dataloader $D\in\{d_{j}\}_{j=0}^{J}$ with $b$ and $s^{hw}$.
    \State Measure training time $t$ of one epoch.
    \For{$d_{j} \gets D$}
        \State $\mathcal{L}=f_{\theta_{j}, q}^{i}(d)$
        \If{$D$ is train set}
            \State $\theta_{j+1, q} \gets \theta_{j, q} - \eta\nabla \mathcal{L}(\theta_{j, q})$
        \EndIf
    \EndFor
    \State \Return $t$
\EndFunction
\State \emph{\textbf{Instantiation phase: } Estimate and sort out the given candidates $C$ through the order of validation accuracy on ImageNet-100 benchmark.}
\For {$f_{\theta, q}^{i},s^{hw}, b^{i}, m^{i} \gets C$}
    \For{$e \gets$ $m^{i}$}
        \State $train\_acc \gets$ TRAIN\_MODEL($f_{\theta, q}^{i}, s^{hw}, b^{i}$)
        \If{$m^{i}-e < 3$}
            \State $val\_acc \gets$ VAL\_MODEL($f_{\theta, q}^{i}, s^{hw}, b^{i}$)
        \EndIf
    \EndFor
\EndFor

\end{algorithmic}
\end{algorithm}

\begin{table*}[]
\centering
\resizebox{\textwidth}{!}{
\begin{tabular}{|c|c|c|c|c|c|c|c|}
\hline
\multicolumn{1}{|c|}{Methods} & Train image-size & Test image-size & Ensemble (En) & AMP & Val Acc. & Phase 1 Test Acc. & Phase 2 Test Acc. \\ \hline
Symmetric-sizes (SS) & 160 & 160 &  &  & 87.08 & 84.62 & 86.28 \\ \hline
Asymmetric-sizes (AS) & 160 & 224 &  &  & 88.70 & 86.40 & 87.60 \\ \hline
AS w/ En & 160 & 224 & \checkmark &  & 90.80 & - & 90.34 \\ \hline
AS w/ AMP & 160 & 224 &  & \checkmark & 91.20 & - & - \\ \hline \hline
Ours (AS w/ En and AMP) & 160 & 224 & \checkmark & \checkmark & \textbf{91.60} & \textbf{89.16} & \textbf{91.38} \\ \hline
\end{tabular}
}
\caption{Quantitative results on our method and ablation studies of key components at training and inference stages. The usage of input images with higher resolution at the inference stage improves 1.4\%p validation accuracy (Val Acc.), and ensemble with the results from image-sizes 160 and 224 helps Val Acc. to rise an extra 2.1\%p from the methods with asymmetric-sizes (AS). Moreover, adopting mixed precision training helps to improve training speed and enlarge batch-size so that the score of methods with AMP goes up about 2.5\%p.
}
\vspace{-1.0ex}
\label{tab1}
\end{table*}

\begin{table}[]
\centering
\resizebox{\columnwidth}{!}{
\begin{tabular}{|c|c|c|c|c||c|}
\hline
Models & \# of params. & Batch-size & Image-size & Max-epochs & Val Acc. \\ \hline
EfficientNet\_B0~\cite{tan2020efficientnet} &  4.1M  & 110 & 160 & 112 & 86.51 \\ \hline
MobileNetV3~\cite{DBLP:journals/corr/abs-1905-02244} &  4.3M  & 200 & 160 & 110 & 84.12 \\ \hline
EfficientNet\_B1~\cite{tan2020efficientnet} &  6.6M  & 80 & 160 & 62 & 85.22 \\ \hline
ResNet18~\cite{DBLP:journals/corr/HeZRS15} &  11.2M  & 256 & 160 & 110 & 83.24 \\ \hline
ResNest26d~\cite{zhang2020resnest} &  15.2M  & 110 & 160 & 51 & 85.71 \\ \hline
ResNet50~\cite{DBLP:journals/corr/HeZRS15} &  23.7M  & 100 & 160 & 60 & 86.92 \\ \hline
\textbf{ResNest50d$^{1}$}~\cite{zhang2020resnest} &  23.8M  & \underline{56} & 160 & \underline{46} & \textbf{87.08} \\ \hline
ResNest50d~\cite{zhang2020resnest} &  25.6M  & 74 & 160 & 46 & 87.02 \\ \hline
ResNest50d$^{2}$~\cite{zhang2020resnest} &  28.5M  & 58 & 160 & 36 & 87.01 \\ \hline
ResNet101~\cite{DBLP:journals/corr/HeZRS15} &  42.7M  & 64 & 160 & 38 & 84.71 \\ \hline
\end{tabular}
}
\caption{Accuracy of candidate models trained with varying batch-sizes, image-sizes, and max-epochs. Val Acc. represents the accuracy of the validation dataset. ResNest50d$^{1}$ and ResNest50d$^{2}$ denote ResNest50d\_1s4x24d and ResNest50d\_4s2x40d, respectively.}
\vspace{-1.0ex}
\label{tab2}
\end{table}

\subsection{Implementation Details}
We adjust the time-limitation for training models from 9 to 3 hours on RTX 3090~\cite{rtx}, adopt ResNest50d\_1s4x24d ~\cite{zhang2020resnest} as our backbone, and set batch-size and max-epochs 65 and 46 in each. In addition, we employ the AdamW optimizer ~\cite{loshchilov2019decoupled} and cosine scheduler ~\cite{loshchilov2017sgdr}.

\subsection{Enlargement of batch-size using AMP}
Our model is trained with mixed precision for less memory usage. As the required memory budget decreased by using AMP, it is possible \textit{to increase the batch-size} \underline{56 to 96}. In addition, this leads to acceleration of the training speed so that \textit{maximum epochs stretch} \underline{46 to 72}. As shown in Table \ref{tab1}, our method with mixed precision exceeds the model with no-AMP by 3\%p higher validation accuracy. 
When the learnable-parameters of our model are calculated in a half-precision floating point format, the throughput becomes higher. Mixing FP16 and FP32 is automatically calculated by PyTorch in this challenge.

\subsection{Asymmetric training and deploying image-sizes}
To enforce the GPU memory constraint, we utilize the asymmetric image-sizes 160 and 224 for training and deploying individually. Moreover, our multi-inference ensembles combine the outputs of our model according to the test images and arbitrarily flipped test images. As indicated in Table ~\ref{tab1}, our approaches demonstrate consistently improved performance because the high-resolution images contain abundant information and the flipped images enable the randomness of our trained model.

\subsection{Resource-aware backbone exploration}
As illustrated in Table ~\ref{tab2}, we present candidate models with maximize batch-size and training epochs by our Algorithm ~\ref{alg:cap1}. Those confirmed-parameters lead to an adaptive learning-rate for learning our models with a cosine annealing scheduler ~\cite{loshchilov2017sgdr}. Consequently, we evaluate that ResNest50d$^{1}$(radix 1, cardinality 4, and base-width 24) ~\cite{zhang2020resnest} serves as the most suitable backbone to adapt our methods: 1) increasing batch-size using AMP and 2) employing asymmetric training and deploying image-sizes. Finally, we validate that each of the two components contributed to performance improvement because larger batch-size, image-size, and epochs affect intra-domain performance in general as shown in Table ~\ref{tab1}.

\section{Conclusion}
We draw our approaches for International Conference on Computer Vision (ICCV) 2023 Workshop Challenge Track 1 on \textbf{R}esource Efficient Deep Learning for \textbf{C}omputer \textbf{V}ision (RCV): Budgeted Model Training Challenge. Based on our approaches, we achieved first place in the challenge as indicated (our team named \textbf{\underline{helloimyjk}}) on the final leader-board~\footnote{https://sites.google.com/view/rcv2023/leaderboard-workshop-challenges.}.

\vspace{-1.0ex}
\section*{Acknowledgements}
\vspace{-1.0ex}
This work was supported by KakaoBank Corporation.

{\small
\bibliographystyle{ieee_fullname}
\bibliography{egbib}
}

\end{document}